\documentclass[conference]{IEEEtran}
\IEEEoverridecommandlockouts

\usepackage{cite}
\usepackage{amsmath,amssymb,amsfonts}
\usepackage{algorithmic}
\usepackage{graphicx}
\usepackage{textcomp}
\usepackage{xcolor}
\usepackage{caption}
\usepackage{subcaption}
\usepackage{multirow}
\usepackage{booktabs}

\def\BibTeX{{\rm B\kern-.05em{\sc i\kern-.025em b}\kern-.08em
    T\kern-.1667em\lower.7ex\hbox{E}\kern-.125emX}}
\begin{document}

\title{Multilingual Attribute Extraction \\from News Web Pages}

\author{\IEEEauthorblockN{Pavel Bedrin}
\IEEEauthorblockA{\textit{ISP RAS, MSU} \\
Moscow, Russia \\
pbedrin@ispras.ru}
\and
\IEEEauthorblockN{Maksim Varlamov}
\IEEEauthorblockA{\textit{ISP RAS} \\
Moscow, Russia \\
varlamov@ispras.ru}
\and
\IEEEauthorblockN{Alexander Yatskov}
\IEEEauthorblockA{\textit{ISP RAS, MSU} \\
Moscow, Russia \\
yatskov@ispras.ru}
}

\maketitle

\begin{abstract}
This paper addresses the challenge of automatically extracting attributes from news article web pages across multiple languages.
Recent neural network models have shown high efficacy in extracting information from semi-structured web pages.
However, these models are predominantly applied to domains like e-commerce and are pre-trained using English data, complicating their application to web pages in other languages.
We prepared a multilingual dataset comprising 3,172 marked-up news web pages across six languages (English, German, Russian, Chinese, Korean, and Arabic) from 161 websites. The dataset is publicly available on GitHub\footnote{https://github.com/ispras/news-page-dataset}.
We fine-tuned the pre-trained state-of-the-art model, MarkupLM, to extract news attributes from these pages and evaluated the impact of translating pages into English on extraction quality.
Additionally, we pre-trained another state-of-the-art model, DOM-LM, on multilingual data and fine-tuned it on our dataset.
We compared both fine-tuned models to existing open-source news data extraction tools, achieving superior extraction metrics.
\end{abstract}


\section{Introduction}
Automatic extraction of news web page attributes is necessary to solve various application tasks, such as news aggregation, recommendation systems and media analysis. To accomplish these tasks, it is essential to develop a method that can automatically process a substantial volume of websites in multiple languages.

Recent studies have shown benefits in using neural network transformer models to solve the task of automatic extraction of news web page attributes. Typically, these models initialized from language models and then pre-trained on web pages using various methods. However, this method has a significant drawback: it only works well on pages in the same language as the training data.

We investigated the effectiveness of the MarkupLM \cite{markuplm} model pre-trained on data in English in extracting news website attributes for non-English languages. We trained the multilingual DOM-LM \cite{domlm} model to tackle the problem and achieved the highest performance on most news website attributes among all methods tested.

\section{Related Work}

In this section, we review existing publicly available datasets for structured web data extraction, along with modern solutions for this task, with a particular focus on BERT-like transformer models.

\subsection{Datasets}

There are several publicly available datasets, which provide a rich source of data for researchers engaged in the task of extracting website attributes. Different approaches have unique dataset requirements that are challenging to implement within a single one: diversity of websites, the number of pages for each, storage formats and attribute sets.

One of the most prevalent datasets in the task is SWDE \cite{swde} (2011), which is extensively utilized for evaluating the quality of structured data extraction. It includes 8 subject domains (verticals), each consisting of 10 sites, ranging from 4,405 to 20,000 pages per vertical. The total number of pages in the dataset is 124,291. Each vertical has an average of 4 attributes. SWDE includes HTML files of the pages and attribute data stored separately and not tied to any specific nodes. All websites are in English.

Klarna Product Page Dataset \cite{klarna} (2021) consists of web pages in MHTML format, their screenshots and snapshots\footnote{https://github.com/klarna-incubator/webtraversallibrary}. It includes 8,175 sites with 51,701 e-commerce product pages in total. The sites of 8 languages are presented: DE, US, GB, FI, AT, SE, NO, NL. Each page contains 5 labeled attributes: buy button, cart button, product price, name and image.

The CoVA dataset \cite{cova} (2022) is designed for approaches based on visual page analysis instead of HTML code. It includes web page images and node characteristics: node boundaries, the number of words in the node text, and the presence of special characters. The CoVA dataset comprises 408 e-commerce websites in various languages, totaling 7,740 pages, and includes the following attributes: product price, title, and image.

Researchers utilize datasets with limited page numbers yet diverse websites to assess information extraction method quality. For instance, the Zyte Article Extraction Benchmark\footnote{https://github.com/scrapinghub/article-extraction-benchmark} (2020) contains 181 HTML files from unique websites, their url, screenshot, and article text. The quality of article text extraction is evaluated. The Zyte Product Extraction Benchmark\footnote{https://github.com/scrapinghub/product-extraction-benchmark} (2021) contains 140 products pages from different websites with labeled price, stock-keeping unit (SKU), availability, in-stock status, and out-of-stock status. These benchmarks are used to evaluate both open-source libraries and commercial services, such as Diffbot and Zyte Automatic Extraction.

The article \cite{newsdataset} (2022) describes the creation of a dataset of Russian-language news web pages. It consists of 722 news pages from 112 websites. The pages are annotated with 9 attributes: title, subtitle, publication and modification dates, text, tags, category, author, source. Text, XPath of start and end HTML nodes, start and end text offsets within these nodes, and global offsets in HTML code are provided for each attribute.

Available datasets statistics are summarized in table \ref{table:datasets}.

\begin{table}[ht!]
    \caption{Datasets for information extraction from web pages}
    \begin{center}
    \renewcommand{\arraystretch}{1.15}

\begin{tabular}{| c | c c c c c |}
    \hline
    Dataset & Year & Domain & Sites & Pages & Language\\
    
    \hline
    SWDE & 2011 & Various & 10 & 124,291 & English\\
    Klarna & 2021 & E-commerce & 8175 & 51,701 & Various\\ 
    CoVA & 2022 & E-commerce & 408 & 7,740 & Various\\ 
    Zyte Products & 2021 & E-commerce & 140 & 140 & Various\\
    Zyte Articles & 2020 & Articles & 181 & 181 & Various\\
    Russian News & 2022 & News & 112 & 722 & Russian\\ 
    \hline
\end{tabular}
    \end{center}
    \label{table:datasets}
\end{table}

\subsection{Extraction Methods}
Automated methods allow to extract attributes from any website within a specific domain without having to manually create a data collection algorithm for each site. These methods can be broadly classified into two primary categories: heuristic tools and approaches based on neural network models. We will consider BERT-based transformer models, which are among the most promising approaches in this area.

MarkupLM \cite{markuplm} (2022) is a pre-trained model designed specifically for document understanding tasks that utilize markup languages, such as HTML, where text and markup information are jointly pre-trained. Text is encoded by RoBERTa \cite{roberta} model. XPath expressions of HTML nodes form the relationship between document elements. MarkupLM model was pre-trained on 24M English web pages from CommonCrawl\footnote{https://commoncrawl.org/} dataset with three strategies: Masked Markup Language Modeling, Node Relation Prediction and Title Page Matching. The pre-trained MarkupLM model has been made available for fine-tuning by the developers, with a focus on two specific downstream tasks: information extraction and reading comprehension. The information extraction task is formulated as token classification into $n+1$ classes ($n$ is the number of extracted attributes, additional category is \textit{None}). 

DOM-LM \cite{domlm} (2022) is also RoBERTa-based pre-trained model, but unlike MarkupLM, DOM-LM considers various structural features of nodes instead of relying on XPath expressions. Authors propose the DOM Tree Processor algorithm for splitting DOM tree into subtrees that preserve local context. Concatenation of its HTML tag, attribute markups and textual content, which is called node representation, is encoded by RoBERTa. Node depth and index, parent node index and some other characteristics are used as tree position features. DOM-LM was pre-trained on over 120,000 English web pages from SWDE using an adapted Masked Language Modeling strategy.

Structor \cite{structor} (2023) is based on MarkupLM and incorporates site-level information and attribute patterns by regular expressions for each DOM node. The authors retrive a node in the same position from another DOM tree on the same website and then incorporate it into the input sequence, approximating character patterns of DOM nodes using regular expressions and integrating them into the neural networks logits.

WebLM \cite{weblm} (2024) is a multimodal pre-trained network designed to address the limitations of solely modeling text and structure modalities of HTML. It integrates the hierarchical structure of document images to enhance the understanding of markup-language-based documents, significantly outperforming previous state-of-the-art models on SWDE dataset.

\begin{table}[ht!]
\caption{Extraction performance of existing models on SWDE (F1-score)}
\label{table:existing_models_compare}
\begin{center}
\renewcommand{\arraystretch}{1.15}

\begin{tabular}{|l|cccc|}
\hline
& MarkupLM & DOM-LM & Structor & WebLM \\ 
\hline
\multicolumn{1}{|l|}{$k = 2$} & 91.29        & 91.70      & 92.12        & 93.17     \\ 
\multicolumn{1}{|l|}{$k = 5$} & 95.89        & 95.70      & 96.08        & 96.78     \\ 
\hline
\end{tabular}
\end{center}
\end{table}

Table \ref{table:existing_models_compare} shows comparative testing results of some models on SWDE with metrics from related papers. For each SWDE vertical, the models were fine-tuned on $k=2$ and $k=5$ random sites and then evaluated on the remaining $10 - k$ sites. Ten training samples were constructed by cyclically shifting the list of sites.

\section{Problem Formulation}

In this paper, we discuss methods for automatically extracting news articles from web pages in various languages. Given an HTML page containing a single news article, the method must extract its attributes such as the title, publication date, main text, authors, and tags, returning them in a structured format (e.g., JSON). The method must be applicable to new websites that were not encountered during its training. Proposed solutions must be evaluated on a dataset containing articles in a diverse set of languages from different language groups.

\section{Dataset Construction}

There is already a dataset of labeled news articles in Russian, described in \cite{newsdataset}.
We decided to enrich it with other languages and collected news websites in English, German, Chinese, Korean, and Arabic.

\begin{figure}[t]
     \centering
     \begin{subfigure}[b]{0.48\textwidth}
         \centering
         \includegraphics[width=\textwidth]{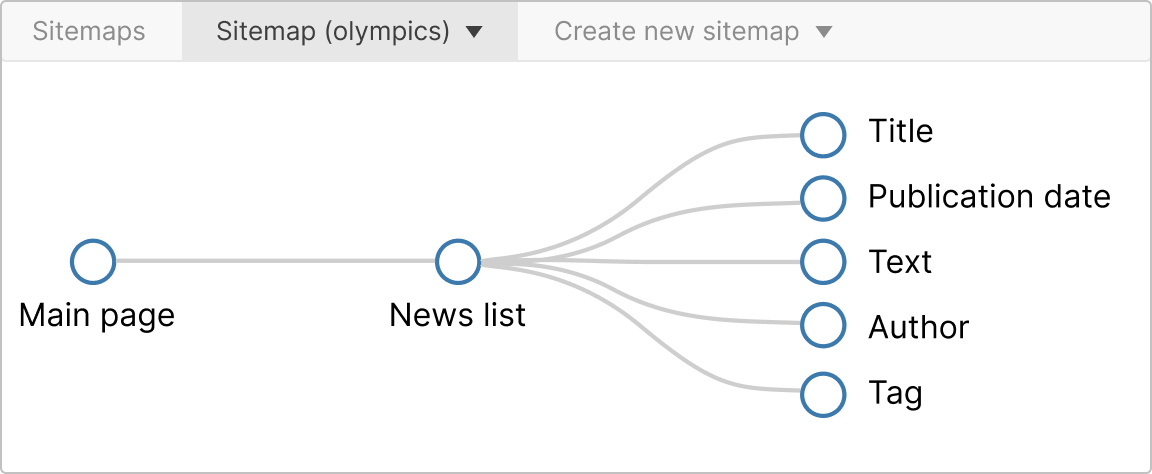}
         \caption{Sitemap graph}
         \label{figures:webscraper-graph}
         \vspace*{3mm}
     \end{subfigure}
     \begin{subfigure}[b]{0.475\textwidth}
         \centering
         \includegraphics[width=\textwidth]{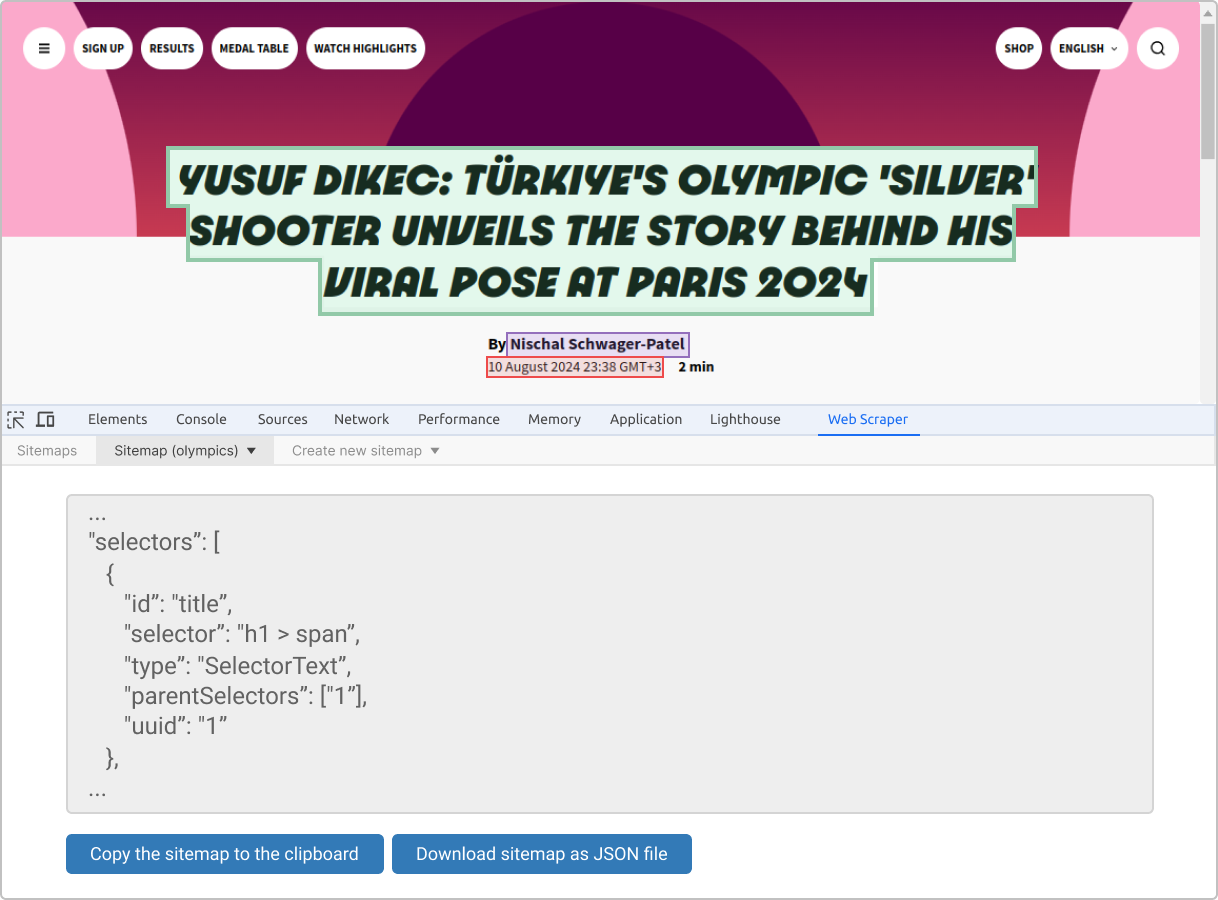}
         \caption{Sitemap fragment}
         \label{figures:webscraper-json}
     \end{subfigure}
        \caption{Sitemap example in Web Scraper}
        \label{figures:webscraper}
\end{figure}

The labeling of the Russian dataset was performed manually by a team of five annotators using the Label Studio tool \cite{label-studio}. This page-by-page markup showed high inter-annotator agreement, but it requires an extremely long time and labor for large amounts of data. To mark up pages in new languages, we used Web Scraper\footnote{https://github.com/ispras/web-scraper-chrome-extension} utility. It allows to interactively create website wrappers, called sitemaps, describing the order of resource traversal and CSS selectors of attributes to be extracted from pages. An example of creating a sitemap in the Web Scraper GUI is shown in Figure \ref{figures:webscraper}.

Our team of annotators manually created sitemaps for each website. We labeled the 5 most common and important attributes: title, publication date, text, author, and tags. The annotators followed the rules of the guideline that we compiled. The key rules were as follows:
\begin{itemize}
    \item Attribute selectors should not locate unnecessary text.
    \item It is allowed to select multiple entities on the page for all attributes except title and publication date.
    \item If some attribute has a variable location on different site pages, so we cannot write a universal selector for it, we do not label this attribute.
\end{itemize}

A special crawler based on Scrapy\footnote{https://scrapy.org/} was used to collect HTML pages and extract article data using created sitemaps. We used Puppeteer\footnote{https://pptr.dev/} to render pages and saved them in HTML and MHTML formats.

After that, we did a multi-step work on data validation and error correction. We tried to extract attribute nodes from downloaded pages using sitemap selectors and collected cases when no nodes were found or the extracted data did not meet our heuristic requirements. We manually checked these problem cases and fixed them. Firstly, we resolved the issues related to page loading:
\begin{itemize}
    \item If the page did not load completely, we retried the download.
    \item If the site was denied access due to regional restrictions, we used a set of proxies from different countries to download it.
    \item If the site contained a captcha that can be bypassed by just clicking, we re-downloaded such pages with an automatic captcha clicking.
    \item Sites with more complex captchas, offline sites, redirected or not found pages were skipped.
\end{itemize}
Next, we fixed problems with the correctness of selectors:
\begin{itemize}
    \item If the site layout changed after the sitemap creation or the selector located nothing, we corrected it.
    \item If more than one node was found for the title or publication date, we corrected the selector.
    \item Annotators were able to select node attribute values instead of node text in Web Scraper. Since our task uses textual features, we remade such attribute-value selectors.
\end{itemize}

For each page, we stored its URL, HTML and textual attribute values in JSON format. Each page contained exactly one article. We built DOM trees for pages to get XPath of attributes by their CSS selectors since XPath is used by some models.

In total, we have prepared 3,172 pages with news articles from 161 website, attempting to balance number of pages for each language. The complete dataset statistics are presented in Table \ref{table:multids_stats}.

\begin{table}[t]
\caption{Attribute statistics in a multilingual dataset}
\begin{center}
\renewcommand{\arraystretch}{1.15}

\begin{tabular}{|c|c|ccccc|}
\hline
   &&
   \multicolumn{1}{c}{Title} &
   \multicolumn{1}{c}{Text} &
   \multicolumn{1}{c}{Date} &
   \multicolumn{1}{c}{Author} &
   \multicolumn{1}{c|}{Tag}
\\
\hline
\multirow{4}{*}{en}
& Sites / Pages
& \multicolumn{5}{c|}{10 / 500} \\
\cline{2-7}
& Sites with attr
& 10 & 10 & 10 & 4 & 2 \\

& Pages with attr
& 500 & 499 & 499 & 147 & 98 \\

& Nodes with attr
& 500 & 22200 & 499 & 147 & 258 \\

\cline{1-7}
\multirow{4}{*}{de}
& Sites / Pages
& \multicolumn{5}{c|}{9 / 450} \\
\cline{2-7}
& Sites with attr
& 9 & 9 & 9 & 9 & 2 \\

& Pages with attr
& 450 & 449 & 450 & 270 & 100 \\

& Nodes with attr
& 454 & 6847 & 600 & 308 & 336 \\

\cline{1-7}
\multirow{4}{*}{ru}
& Sites / Pages
& \multicolumn{5}{c|}{112 / 722} \\
\cline{2-7}
& Sites with attr
& 110 & 112 & 110 & 54 & 49 \\

& Pages with attr
& 712 & 716 & 708 & 262 & 332 \\

& Nodes with attr
& 714 & 5918 & 724 & 272 & 1190 \\

\cline{1-7}
\multirow{4}{*}{zh}
& Sites / Pages
& \multicolumn{5}{c|}{10 / 500} \\
\cline{2-7}
& Sites with attr
& 10 & 10 & 10 & 6 & 0 \\

& Pages with attr
& 500 & 500 & 500 & 227 & 0 \\

& Nodes with attr
& 501 & 5872 & 500 & 277 & 0 \\

\cline{1-7}
\multirow{4}{*}{ko}
& Sites / Pages
& \multicolumn{5}{c|}{10 / 500} \\
\cline{2-7}
& Sites with attr
& 10 & 10 & 10 & 8 & 1 \\

& Pages with attr
& 500 & 500 & 500 & 358 & 41 \\

& Nodes with attr
& 500 & 6898 & 550 & 409 & 155 \\

\cline{1-7}
\multirow{4}{*}{ar}
& Sites / Pages
& \multicolumn{5}{c|}{10 / 500} \\
\cline{2-7}
& Sites with attr
& 10 & 10 & 10 & 5 & 4 \\

& Pages with attr
& 500 & 500 & 500 & 180 & 184 \\

& Nodes with attr
& 500 & 5752 & 550 & 274 & 648 \\

\hline
\end{tabular}
\label{table:multids_stats}
\end{center}
\end{table}

\section{English language Model}

During fine-tuning, the authors of the MarkupLM \cite{markuplm} paper published a pre-trained model\footnote{https://huggingface.co/docs/transformers/model\_doc/markuplm} on English data. We want to experimentally evaluate the quality of MarkupLM when handling non-English news pages.

We preprocessed our dataset for MarkupLM using the methodology proposed by its authors. Firstly, we built a DOM tree using HTML and cleaned it of non-text nodes. Then we cleaned the texts of unnecessary whitespace and control characters. We mapped the attributes to the corresponding nodes in the DOM tree using our labeled dataset. As a result, for each page we have a list of tuples like \texttt{\{node text, node XPath, node label\}}.

All text nodes were tokenized and each token was mapped to a label, a sequence of tokenized XPath tags and subscripts of its node. For the model evaluation phase, we stored the page ID and node text.

Pages were divided into chunks (that is, sequences of length \texttt{max\_seq\_length}) intersecting with fixed step \texttt{doc\_stride}.  We set \texttt{max\_seq\_length} to 512 as the maximum length of MarkupLM input, \texttt{doc\_stride} was set to 170 as 1/3 of this length.

\section{Multilingual Model}

Modern BERT-like models can be made multilingual. To do this, we need to initialize the model from a multilingual language model, use a multilingual tokenizer, and train on multilingual data.

We chose DOM-LM model for this task. It showed high, close to MarkupLM results on SWDE, but requires several times less pre-training data volume. We adapted the implementation\footnote{https://github.com/Misterion777/DOM-LM} of DOM-LM pre-trained on SWDE for our multilingual scenario.

We also utilized a multilingual model XLM-RoBERTa \cite{xlmroberta} to evaluate the extraction quality without using the structural markup features. Page preprocessing and chunking was similar to MarkupLM.

\subsection{Data Preprocessing}

The cleaned DOM tree is split into subtrees preserving local context using the author's algorithm. A list of subtree nodes is generated by pre-order traversal. Tokenized concatenation of tag, attributes and text, which is called textual features, and structural features are obtained for each node.

Additionally, for fine-tuning we set the node label to the BOS-token and ignore index for PyTorch, $-100$, to the remaining tokens. BOS-token is also considered for obtaining node prediction. 

\subsection{Prediction Aggregation Strategy}

The DOM-LM model solves token classification task, i.e. each token of a page text has its own prediction. To move from token classification to node classification, we need to combine the predictions of node tokens and get the final node prediction.

\begin{itemize}
    \item In MarkupLM labels are mapped to all node tokens in the training data and the first token is used for prediction.
    \item The DOM-LM architecture scheme indicates that the label is set on the node BOS token. Prediction is also taken from it. Each node starts with a BOS token and ends with an EOS token.
    \item Some other aggregation strategies for the NER task are described in HuggingFace Transformers documentation\footnote{https://huggingface.co/docs/transformers/v4.15.0/en/main\_classes/pipelines}.
\end{itemize}

We have implemented several aggregation strategies:
\begin{itemize}
    \item \textbf{BOS}. We made it according to the DOM-LM paper.
    \item \textbf{FIRST}. Using training data, labels on all node tokens. Prediction is based on the first token.
    \item \textbf{AVG}. Using training data, labels on all node tokens. Prediction is averaging probabilities over tokens.
    \item \textbf{MAX}. Using training data, labels on all node tokens. Prediction is based on the token with the highest confidence in label.
    \item \textbf{ANY}. Using training data, labels on all node tokens. Among tokens with predictions other than \textit{None}, we select token with the highest confidence in label.
\end{itemize}

Two versions of the dataset were preprocessed. Class labels were assigned to BOS tokens for the BOS strategy. For other strategies, the label was assigned to all tokens except BOS and EOS. In both variants, the remaining tokens were assigned $-100$.

We evaluated the aggregation strategy impact on quality metrics. Experiment results for Russian and Chinese are presented in Table \ref{table:agg_strategies}. This experiment was performed on a model pre-trained only on our dataset.

From the results, we can conclude that the choice of strategy does not significantly affect the metrics. There is no strategy with consistently higher metrics in most experiments. Therefore, we decided to use the BOS strategy in our experiments, since it is used in DOM-LM paper.

\begin{table}[ht!]
\caption{The impact of prediction aggregation strategy on DOM-LM quality (F1-score)}
\begin{center}
\renewcommand{\arraystretch}{1.15}

\begin{tabular}{|c|l|ccccc|}
\hline
Language                & Strategy & Title & Date & Text & Author & Tag \\ \hline
\multirow{5}{*}{ru} & BOS & 0.96 & 0.95 & 0.93 & 0.59 & 0.76 \\
                    & FIRST     & 0.96 & 0.93 & 0.94 & 0.51 & 0.78 \\
                    & AVG       & 0.96 & 0.93 & 0.93 & 0.51 & 0.78      \\
                    & MAX       & 0.96 & 0.92 & 0.93 & 0.49 & 0.75      \\
                    & ANY       & 0.96 & 0.94 & 0.93 & 0.50 & 0.77      \\
                    \hline
\multirow{5}{*}{zh} & BOS       & 0.70 & 0.74 & 0.84 & 0.00 & -      \\
                    & FIRST     & 0.67 & 0.71 & 0.84 & 0.03 & -      \\
                    & AVG       & 0.67 & 0.75 & 0.84 & 0.01 & -      \\
                    & MAX       & 0.66 & 0.72 & 0.83 & 0.01 & -      \\
                    & ANY       & 0.68 & 0.74 & 0.84 & 0.06 & -      \\ \hline
\end{tabular}
\label{table:agg_strategies}
\end{center}
\end{table}

\subsection{Model Pre-training}

To obtain a multilingual DOM-LM we first initialize its weights from the multilingual XLM-RoBERTa \cite{xlmroberta} instead of the English-language RoBERTa model as in the original paper. XLM-RoBERTa was trained on 2.5 TB web pages of 100 languages from the CommonCrawl dataset. 

DOM-LM pre-training was performed on our multilingual news dataset and on a one-day sample of 37,473 news pages of different languages from the CommonCrawl-News dataset\footnote{https://data.commoncrawl.org/crawl-data/CC-NEWS/index.html}. 

\section{Experiment Setup}

In this section, we outline the evaluation scheme for the subsequent experiments and the parameters used for pre-training and fine-tuning the models.

\subsection{Evaluation Metric}

We employ quality assessment methodology from \cite{newsdataset}, which is based on Zyte's benchmarks for article and product data extraction.
For each news page, the attribute values extracted by automatic methods are compared with the ground truth values. Based on this, we computed true positives, false positives and false negatives. Then we use $F_1$-score. Different matching strategies were applied depending on the attributes:
\begin{itemize}
    \item Titles and texts were compared as bags of 4-grams. This makes sense, as individual words can be seen in both correctly extracted text segments and undesirable ones, whereas for n-grams, this is significantly less likely.
    \item Authors and tags were compared as sets of individual values, with lowercase and strip normalization applied.
    \item Dates were compared after parsing and normalization using the dateparser\footnote{https://github.com/scrapinghub/dateparser} library.

\end{itemize}

The metrics are averaged across pages. Sites used for model training and evaluation did not intersected.

The models were assessed using the 5-Fold cross-validation method. The set of websites was divided into 5 parts, where 4 parts were used for training and the remaining part was used for testing. Finally, the results of the 5 experiments were averaged. The website splits remained unchanged between experiments. This method evaluates the model quality with the most uniform data utilization.

\subsection{Settings}

We have pre-trained DOM-LM almost according to the original paper. It was initialized with pre-trained XLM-RoBERTa-Base using Transformers\footnote{https://huggingface.co/FacebookAI/xlm-roberta-base}.Then pre-trained on raw HTML documents without labels for 5 epochs with a batch size of 24, a linear learning rate scheduler with max learning rate of $1e^{-4}$ and warm up for the first half epoch.

For fine-tuning all models were trained for 1 epoch with a batch size of 8, and the AdamW optimizer with a linear scheduler. The scheduler had a learning rate of $3e^{-5}$, weight decay of $0$ and epsilon of $1e^{-8}$.

The models were trained on one NVIDIA A100 80GB GPU. It took approximately 5 days to pre-train DOM-LM.

\section{Experiments}

In this section, we discuss the experiments conducted. We performed several sets of experiments with the described models on our data:
\begin{itemize}
\item fine-tuning and evaluation on the single-language subsets of the dataset
\item fine-tuning on one language and evaluating on another
\item fine-tuning and evaluation on the full dataset with mixed languages
\end{itemize}

Additionally, we evaluated the models and open-source tools on the Zyte Article Extraction Benchmark.

\subsection{One-language Evaluation}

\begin{table}[ht!]
\caption{Methods quality in one-language experiments (F1-score)}
\begin{center}
\begin{tabular}{|l|l|ccccc|}
\hline
 & Method & \multicolumn{1}{l}{Title} & \multicolumn{1}{l}{Date} & \multicolumn{1}{l}{Text} & \multicolumn{1}{l}{Author} & \multicolumn{1}{l|}{Tag} \\
\hline
\multirow{6}{*}{en} & Trafilatura & \textbf{1.00} & 0.49 & 0.86 & \textbf{0.53} & 0.12 \\
                    & Newspaper  & 0.94 & 0.42 & \textbf{0.89} & 0.36 & \textbf{0.89} \\
                    & Newsplease   & 0.95 & 0.50 & \textbf{0.89} & 0.36 & 0.00 \\
                    & XLM-RoBERTa & 0.47 & 0.57 & 0.74 & 0.33 & 0.35 \\
                    & MarkupLM    & 0.94 & 0.56 & 0.66 & 0.00 & 0.36 \\
                    & MarkupLM-EN & 0.94 & 0.56 & 0.66 & 0.00 & 0.36 \\
                    & DOM-LM      & 0.85 & \textbf{0.64} & 0.83 & 0.00 & 0.00 \\ \hline
\multirow{6}{*}{de} & Trafilatura & 0.52 & 0.37 & \textbf{0.89} & 0.24 & \textbf{0.24} \\
                    & Newspaper  & 0.46 & 0.11 & 0.84 & 0.17 & 0.00 \\
                    & Newsplease   & 0.53 & 0.45 & 0.84 & 0.17 & 0.00 \\
                    & XLM-RoBERTa & 0.55 & 0.56 & 0.87 & 0.58 & 0.01 \\
                    & MarkupLM    & 0.74 & 0.52 & 0.56 & 0.47 & 0.00  \\
                    & MarkupLM-EN & 0.78 & \textbf{0.74} & 0.69 & 0.47 & 0.00  \\
                    & DOM-LM      & \textbf{0.83} & 0.72 & 0.87 & \textbf{0.70} & 0.00  \\ \hline
\multirow{6}{*}{ru} & Trafilatura & 0.96 & 0.42 & 0.78 & 0.44 & 0.27  \\
                    & Newspaper  & 0.95 & 0.39 & 0.79 & 0.24 & 0.60  \\
                    & Newsplease   & \textbf{0.98} & 0.45 & 0.68 & 0.24 & 0.00  \\
                    & XLM-RoBERTa & 0.92 & 0.94 & 0.91 & 0.66 & 0.83 \\
                    & MarkupLM    & 0.97 & \textbf{0.95} & \textbf{0.96} & 0.48 & 0.75  \\
                    & MarkupLM-EN & \textbf{0.98} & 0.94 & 0.94 & 0.61 & 0.81  \\
                    & DOM-LM      & \textbf{0.98} & 0.94 & 0.95 & \textbf{0.75} & \textbf{0.91}  \\ \hline                  
\multirow{6}{*}{zh} & Trafilatura & 0.75 & 0.29 & 0.83 & 0.00 & - \\
                    & Newspaper  & 0.82 & 0.00 & \textbf{0.93} & 0.00 & - \\
                    & Newsplease   & 0.87 & 0.00 & 0.10 & 0.00 & - \\
                    & XLM-RoBERTa & \textbf{0.97} & \textbf{0.93} & 0.80 & 0.00 & - \\
                    & MarkupLM    & 0.83 & 0.74 & 0.80 & 0.01 & -  \\
                    & MarkupLM-EN & 0.86 & 0.91 & 0.89 & 0.00 & -  \\
                    & DOM-LM      & 0.87 & 0.87 & 0.89 & 0.01 & -  \\ \hline
\multirow{6}{*}{ko} & Trafilatura & 0.94 & 0.19 & \textbf{0.90} & 0.15 & - \\
                    & Newspaper  & 0.99 & 0.00 & 0.79 & 0.10 & - \\
                    & Newsplease   & \textbf{1.00} & 0.42 & 0.64 & 0.10 & - \\
                    & XLM-RoBERTa & 0.83 & 0.80 & 0.82 & 0.19 & - \\
                    & MarkupLM    & 0.41 & 0.67 & 0.79 & 0.00 & -  \\
                    & MarkupLM-EN & 0.90 & 0.79 & 0.88 & 0.18 & -  \\
                    & DOM-LM      & 0.90 & \textbf{0.85} & 0.86 & \textbf{0.21} & - \\ \hline
\multirow{6}{*}{ar} & Trafilatura & 0.95 & 0.35 & \textbf{0.89} & \textbf{0.35} & 0.24 \\
                    & Newspaper  & \textbf{0.97} & 0.25 & 0.85 & 0.15 & 0.50 \\
                    & Newsplease   & \textbf{0.97} & 0.24 & 0.85 & 0.15 & 0.00 \\
                    & XLM-RoBERTa & 0.56 & 0.53 & 0.81 & 0.00 & 0.02 \\
                    & MarkupLM    & 0.66 & 0.37 & 0.68 & 0.00 & 0.01 \\
                    & MarkupLM-EN & 0.94 & \textbf{0.74} & 0.82 & 0.00 & 0.30 \\
                    & DOM-LM      & 0.60 & 0.43 & 0.87 & 0.00 & \textbf{0.38} \\ \hline
\end{tabular}
\label{table:multids_onelang_methods}
\end{center}
\end{table}

MarkupLM uses a byte-level Byte-Pair Encoding (BBPE) \cite{bbpe} tokenizer. It treats text as bytes instead of characters. Therefore, despite the small size of the base dictionary (256 bytes), any characters will be encoded as a byte sequence, without conversion to an out-of-vocabulary token. 

We hypothesized that HTML markup features and the ability to encode any characters would allow MarkupLM to solve the extraction task on multilingual data despite monolingual pre-training data. We evaluated the fine-tuned MarkupLM within each language by conducting experiments in two variants: using original pages and pages translated into English. For the latter, we translated each node in the cleaned HTML pages into English with the Argos Translate\footnote{https://github.com/argosopentech/argos-translate} library.

The experimental results are shown in Table \ref{table:multids_onelang_methods}. 
The model without preliminary translation is denoted as MarkupLM, with translation -- as MarkupLM-EN.
For most languages and attributes, translation improves the extraction quality of MarkupLM. Model without translation does not show consistently high quality on many experiments, it seems that markup information alone is insufficient.

Then we fine-tuned the multilingual DOM-LM. For most languages and attributes, it is close to or better than MarkupLM when using the original pages without translation. 

DOM-LM has room for further improvement. We think that balancing languages in the pre-training data and increasing the diversity of labeled sites will help improve quality. This is particularly evidenced by the higher results for Russian. This dataset has 112 labeled sites instead of 9-10 in other languages. However, the quality of DOM-LM is already higher than MarkupLM or at the same level in most languages.

We also evaluated some open source libraries: Trafilatura\footnote{https://github.com/adbar/trafilatura} \cite{barbaresi-2021-trafilatura} (version 1.11.0), Newspaper\footnote{https://github.com/codelucas/newspaper} (version 0.2.8) and Newsplease\footnote{https://github.com/fhamborg/news-please} \cite{Hamborg2017} (version 1.6.10).
These tools support the extraction of article text and metadata in multiple languages, leveraging semantic markup, various rules, and heuristics.

The evaluation results are shown in Table \ref{table:multids_onelang_methods}. 
The tools exhibit low quality on average when extracting dates, authors, and tags. There is also a noticeable decrease in precision for titles in German, because libraries usually extract the \texttt{<title>} content, but in German media it often includes a \textit{kicker} (short phrase to increase interest).

The overall results of this experiment show that there is no clear leader. Models in most languages do not have enough diversity of training data within languages. The quality of libraries is not stable.

\subsection{New-language Evaluation}

We evaluated the monolingual MarkupLM on an unseen language. The websites of all languages in our dataset, excluding the target, were utilized for fine-tuning. We used pages without translation into english in this experiment. The results are presented in Table \ref{table:except_train}. Notably, the extraction quality for Chinese was higher than for German, although there was no data of the same language group or family as Chinese in the training set. This suggests that the model is more focused on markup data than text when dealing with an unseen language.

\begin{table}[ht!]
\caption{MarkupLM quality in a language not used in fine-tuning (F1-score)}
\begin{center}
\renewcommand{\arraystretch}{1.15}

\begin{tabular}{|l|ccccc|}
\hline
Language & Title & Date & Text & Author & Tag \\
\hline
de & 0.42 & 0.71 & 0.80 & 0.00 & 0.57 \\
zh & 0.84 & 0.86 & 0.90 & 0.00 & - \\
\hline
\end{tabular}

\label{table:except_train}
\end{center}
\end{table}

\subsection{Multi-language Evaluation}

We fine-tuned and evaluated models and libraries on a mixed language dataset. This experiment demonstrates the quality of models in practice when the language of the extracted page can be arbitrary, and the model is trained on all available data. The results of 5-fold cross-validation (all sites in the dataset were randomly split into folds) are presented in Table \ref{table:multilang_methods}.

MarkupLM again demonstrates higher quality when pages are translated into English. The multilingual DOM-LM outperforms other methods in most attributes, while in other cases its results are close to MarkupLM, but DOM-LM does not require the time and computational resources needed for translating. The libraries exhibit lower quality compared to the models.

\begin{table}[ht!]
\caption{Methods quality on multilingual data (F1-score)}
\begin{center}
\renewcommand{\arraystretch}{1.15}

\begin{tabular}{|l|ccccc|}
\hline
Method & Title & Date & Text & Author & Tag \\ 
\hline
Trafilatura & 0.88 & 0.36 & 0.85 & 0.29 & 0.18 \\
Newspaper & 0.89 & 0.21 & 0.72 & 0.17 & 0.55 \\
Newsplease & 0.91 & 0.35 & 0.70 & 0.17 & 0.00 \\
\hline
XLM-RoBERTa & 0.91 & 0.78 & 0.86 & 0.38 & 0.55 \\
MarkupLM & 0.90 & 0.79 & 0.90 & 0.30 & 0.47 \\
MarkupLM-EN & \textbf{0.95} & \textbf{0.88} & 0.91 & 0.40 & 0.67 \\
DOM-LM & 0.93 & 0.87 & \textbf{0.93} & \textbf{0.41} & \textbf{0.73} \\
\hline
\end{tabular}
\label{table:multilang_methods}
\end{center}
\end{table}

\subsection{Zyte Articles Benchmark}

We evaluated open-source libraries and fine-tuned models on Zyte Article Extraction Benchmark dataset. The benchmark task is to extract text from article pages. The dataset contains 181 pages with articles from unique websites of different languages, mostly in English. The evaluation results are shown in Table \ref{table:multids_zyte}. The metrics of proprietary services such as Zyte Automatic Extraction (Zyte AE) and Diffbot are taken from Zyte's whitepaper and also presented.

All methods exhibit a high $F_1$-score. Most model errors occur on non-news articles (e.g., blogs, forums). These pages often contain long comments that the model treats as text, atypically many subheadings, unfamiliar elements, for example, tables that are labeled as text in ground truth.

\begin{table}[ht!]
\caption{Methods quality on Zyte AE Benchmark}
\begin{center}
\renewcommand{\arraystretch}{1.15}

\begin{tabular}{|l|ccc|}
\hline
 Method & \makebox[1.5cm]{F1} & \makebox[1.5cm]{Precision} & \makebox[1.5cm]{Recall} \\
\hline
Zyte AE & 0.97 & 0.98 & 0.96 \\
Diffbot & 0.95 & 0.96 & 0.94 \\
\hline
Trafilatura & 0.95 & 0.92 & 0.97 \\
Newspaper3k & 0.93 & 0.94 & 0.91\\   
Newsplease & 0.92 & 0.94 & 0.91 \\
\hline
XLM-RoBERTa & 0.87 & 0.96 & 0.80 \\
MarkupLM & 0.89 & 0.96 & 0.83 \\
MarkupLM-EN & 0.95 & 0.96 & 0.94 \\
DOM-LM & 0.94 & 0.95 & 0.93 \\
\hline
\end{tabular}
\label{table:multids_zyte}
\end{center}
\end{table}

\section{Conclusions}

In this paper we applied several state-of-the-art models to solve the problem of automatic attribute extraction from multilingual news web pages.

We extended the existing dataset of labeled Russian-language news web pages by adding articles in five new languages (English, German, Arabic, Chinese, and Korean), resulting in a total of 3,172 articles from 161 websites. Extracted attributes are title, publication date, text, authors, and tags.

We adapted the pre-trained English MarkupLM model to our dataset, fine-tuned it on HTML node classification task, and evaluated quality within each language. We found that translating pages into English improves the extraction quality. MarkupLM fails to demonstrate consistently high quality without translation, as markup information alone is insufficient.

We adapted and pre-trained DOM-LM model on news web pages of various languages. The resulting multilingual model was fine-tuned on HTML node classification task. We found that within languages, DOM-LM is comparable or superior to MarkupLM in most experiments, with no data translation overhead. When fine-tuning and evaluating on a mixed language data (a practical use case), multilingual DOM-LM outperforms other solutions for most attributes.

The multilingual DOM-LM has room for further quality improvement through increasing the volume and linguistic balance of pre-training data and a larger number of labeled websites across languages.

\bibliography{main}
\bibliographystyle{IEEEtran}

\end{document}